\documentclass{article}
\usepackage{spconf,amsmath,graphicx,hyperref, bm, amsfonts, makecell}

\title{VNODE: A Piecewise Continuous Volterra Neural Network}
\name{Siddharth Roheda, Aniruddha Bala, Rohit Chowdhury, Rohan Jaiswal}
\address{Samsung Research Institute \\ Bangalore, India \\ \{sid.roheda, aniruddha.b, rohit.c, r.jaiswal\}@samsung.com}
\begin{document}
%
\maketitle
\begin{abstract}

This paper introduces Volterra Neural Ordinary Differential Equations (VNODE), a piecewise continuous Volterra Neural Network that integrates nonlinear Volterra filtering with continuous‑time neural ordinary differential equations for image classification. Drawing inspiration from the visual cortex, where discrete event processing is interleaved with continuous integration, VNODE alternates between discrete Volterra feature extraction and ODE‑driven state evolution. This hybrid formulation captures complex patterns while requiring substantially fewer parameters than conventional deep architectures. VNODE consistently outperforms state-of-the-art models with improved computational complexity as exemplified on benchmark datasets like CIFAR-10 and Imagenet-1K.

\end{abstract}
\section{Introduction}
\label{sec:intro}

Over the past decade, deep learning has transformed signal and image processing. Driven by Convolutional Neural Networks, Transformers, and their variants, it has set benchmarks in image classification, action recognition, object detection, and many other computer vision tasks \cite{zhao2024review}. However, this performance comes at the cost of high memory and computational demands.
Recently, Volterra Neural Networks (VNNs) \cite{roheda2024volterra} were introduced to alleviate this issue. VNNs are based on the Volterra series, which introduces non-linearity through higher-order convolutions, bearing resemblance to information processing observed in the brain's visual system. The mammalian visual cortex, particularly the early visual areas (V1 \& V2), extract and integrate higher order, multiplicative and polynomial interactions among neurons \cite{yu2011higher}. Similarly, VNNs extract features through higher-order convolutions which lead to multiplicative interactions within the signal being processed, and create a polynomial input-output relationship. Prior work \cite{roheda2020conquering},\cite{roheda2024mr} shows that this characteristic of VNNs enables them to out-perform traditional CNN/Transformer based architectures while reducing the computational costs. Despite this strong parallel, it is important to recognize a fundamental difference between them. Traditional VNNs, and other deep architectures, operate as discrete systems with static non-linear layers. In contrast, neurophysiological evidence shows that the brain processes visual information by alternating discrete events (eg. neural spikes in response to incoming stimuli) with phases of continuous, dynamic integration \cite{london2005dendritic}. 

To bridge this gap, we introduce VNODE, embedding Volterra based nonlinear feature extraction within a piecewise continuous neural ODE framework. The proposed model alternates between a higher-order feature extraction layer and a smooth, continuous time evolution of its internal representations. Experiments on image classification show that this design enables parameter efficient modeling of complex patterns without compromising on classification accuracy.

\begin{figure}
    \centering
    \includegraphics[width=0.8\linewidth]{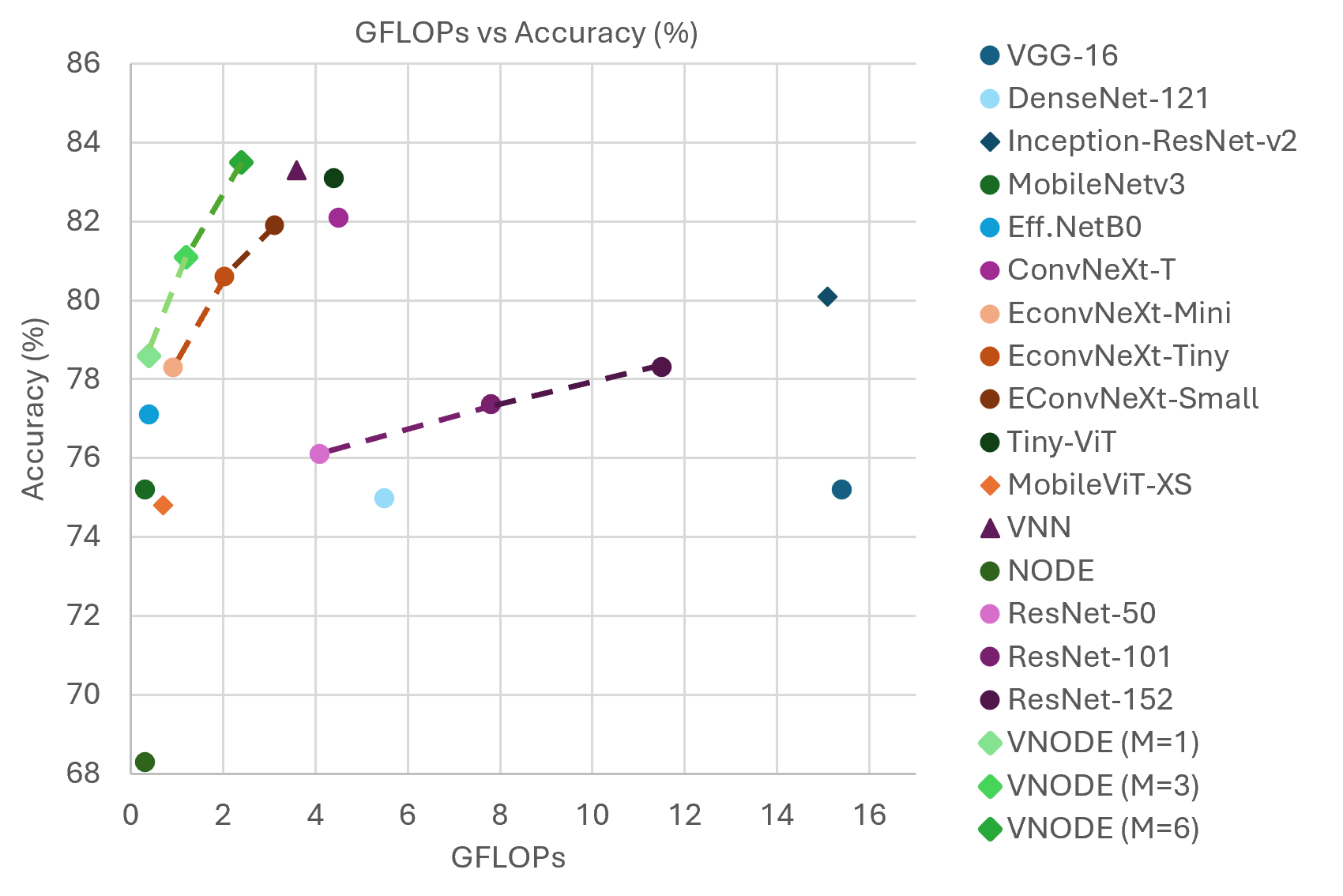}
    \caption{Comparison of Computational Complexity (GFLOPs) and Accuracy (\%) on ImageNet-1K}
    \label{fig:gflopsvsacc}
\end{figure}

\section{Related Work}
\label{sec:related_work}
\subsection{Volterra Filtering in Deep Learning}

The Volterra filter \cite{volterra1930theory} approximates the nonlinear relationship between input $x(t)$ and output $y(t)$ as:
\begin{equation}
    y(t) = b + \sum_{k=1}^K \Bigg[ \sum_{\tau_1=0}^{L-1}...\sum_{\tau_k}^{L-1} \bm{W^k}(\tau_1,...,\tau_k) \prod_{j=1}^k x(t-\tau_j) \Bigg],  
\end{equation}
where $b$ is a bias, $\bm{W^k}$ denotes the $k^{th}$ order Volterra kernel, $K$ is the filter order, and $L$ is its memory. For images, a 2D Volterra filter generalizes this idea:
\begin{align}
    \nonumber y_{\left[\substack{m \\  n }\right]} = b &+ \sum_{k_1,k_2} \bm{W}^1_{\left[\substack{k_1 \\k_2}\right]} x_{\left[\substack{m-k_1 \\ n-k_2}\right]} \\ &+ \sum_{\substack{k_1,k_2 \\ l_1,l_2}} \bm{W}^2 _{\left[\substack{k_1\\k_2}\right] \left[\substack{l_1 \\ l_2}\right]} \bigg[x_{\left[\substack{m-k_1 \\ n-k_2}\right]} \cdot x_{\left[\substack{m-l_1 \\ n-l_2}\right]}\bigg],
\end{align}
where $x_{\left[\substack{i\\j}\right]}$ indexes the input, $y_{\left[\substack{m \\  n }\right]}$ the output, $\bm{W}^k$ is the $k^{th}$ order filter, and $b$ is the bias.
In practice \cite{kumar2011trainable},\cite{gao2019global}, Volterra Filters were limited to second order implementations due to exponential increase in complexity of higher order filters.
Later research \cite{roheda2024volterra},\cite{roheda2020conquering} proposed VNNs where second order Volterra filters were cascaded for modeling higher order non-linearity. In \cite{yu2011higher}, authors note that the human brain’s visual cortex extracts and integrates higher-order multiplicative and polynomial interactions among neurons, hence justifying the use of higher order convolutions in VNNs. Implementations with residual connections \cite{roheda2024mr} showed higher order filters can be trained effectively without loss of performance. VNNs have since demonstrated efficacy in complex non-linear tasks such as action recognition \cite{roheda2020conquering}, media restoration \cite{roheda2024mr}, noise cancellation \cite{bai2025wavenet}, image generation \cite{roheda2024volterra}, and image editing \cite{bala2024galaxyedit}. 

\subsection{Neural ODEs}

Neural Ordinary Differential Equations (NODEs) \cite{chen2018neural} bridge deep learning and dynamical systems, showing that architectures like ResNets \cite{he2016deep} can be viewed as discretizations of continuous transformation flows defined by differential equations. For instance, a ResNet layer is implemented as,
\begin{equation}
    \bm{h}(t+1) = \bm{h}(t) + \bm{f}(\bm{h}(t), \bm{\theta}_t),
\end{equation}
where $t \in \{0,...,T\}$, $\bm{h}(t) \in \mathbb{R}^D$, and $\bm{f}(\cdot)$ is the network layer. This can be interpreted as the Euler discretization of a continuous transformation \cite{haber2017stable, ruthotto2020deep} where the hidden units evolve continuously according to an ordinary differential equation,
\begin{equation}
    \frac{d\bm{h}(t)}{dt} = \bm{f}(\bm{h}(t), t, \theta),
    \label{eq:node}
\end{equation}
where $\bm{f}(\cdot)$ is a non-linear mapping approximated by a neural network. Starting from layer $\bm{h}(0)$, the output layer $\bm{h}(T)$ is defined as the solution to the ODE initial value problem at some time $T$. NODEs have been implemented for regression and classification by mapping input data $\bm{x} \in \mathbb{R}^d$ to features or representations $\bm{\phi}(\bm{x}) \in \mathbb{R}^d$ followed by a linear layer, $\bm{c}: \mathbb{R}^d \to \mathbb{R}$. The features $\bm{\phi}(\bm{x}) = \bm{h}(T)$ are obtained by solving the initial value problem for Equation \ref{eq:node} when $\bm{h}(0) = \bm{x}$. 
NODEs require only $\mathcal{O}$$(1)$ memory and $\mathcal{O}(L)$ computational time, where $L$ is the number of function evaluations of the ODE solver \cite{chen2018neural}. This design enables NODEs to achieve high efficiency in memory usage, parameterization, and computational cost. Since their inception, many extensions have emerged including Augmented NODEs \cite{dupont2019augmented} that improve expressivity by increasing latent dimension, and hybrids like ODE-RNN \cite{habiba2020neural} that integrate ODEs with recurrent architectures for sequence modeling. 

\section{Proposed Approach}

\begin{figure*}
\centering
    \includegraphics[width=0.71\textwidth]{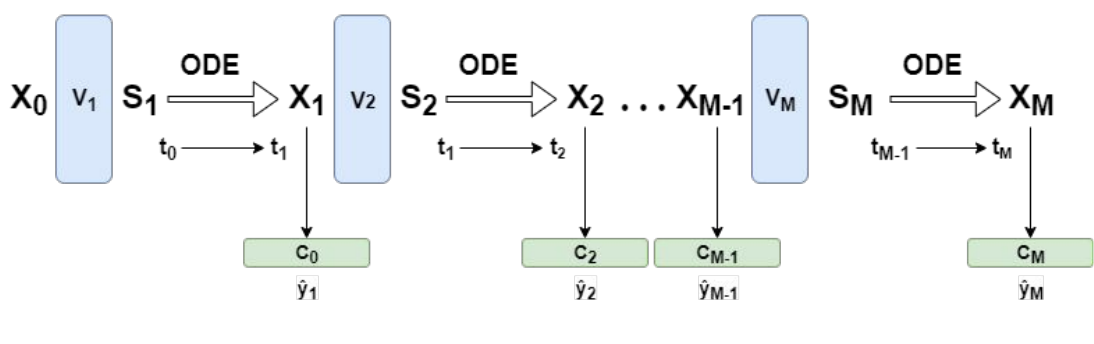}
    \caption{Block Diagram of the Piecewise Continuous VNN}
    \label{fig:bd}

\end{figure*}

In this section, we introduce the piecewise continuous implementation of the Volterra Neural Network called VNODE. Our method leverages the Volterra Filter as the feature extractor and uses them as the building block of an ODE.  



\subsection{Piecewise Continuous VNN}
As discussed in Section \ref{sec:intro}, neurophysiological evidence suggests that the brain processes information by alternating discrete events with phases of continuous, dynamic integration \cite{london2005dendritic}. To this end, we propose a piecewise continuous framework that mimics the human brain’s approach to process information. Let $\bm{X}$ be the input image and define discrete “event" points $t_0 = 0 < t_1 <t_2 < ... < t_M = 1$. At any stage $0 \leq m \leq M$, we perform the following steps: 

\begin{enumerate}
    \item \textbf{Discrete Feature Extraction}: A truncated $K_m^{th}$ order Volterra Filter is used to extract features from the previous stage's output, 
    \begin{equation}
        \bm{S_m} = \bm{V}_m(\bm{X}_{m-1}) = \sum_{k=1}^{K_m} \bm{V}_m^k (\bm{X}_{m-1}),
    \end{equation}

where, $\bm{X}_{m-1}$ is the output of the previous stage, and $\bm{V}_m$ is the discrete Volterra filter at stage $m$ and $\bm{V}_m^k$ is the volterra kernel for the $k^{th}$ order.  



\item \textbf{Continuous Feature Evolution (ODE Block)}: The output of the discrete volterra filter, $\bm{S}_m$ is used as the initial condition for a neural ODE block, $\bm{h}(t_{m-1}) = \bm{S}_m$ in the interval $t_{m-1} \to t_m$,

\begin{equation}
    \frac{d\bm{h}(t)}{dt} = \bm{g}_m (\bm{h}(t), t, \bm{\theta}_m); \text{ } t \in [t_{m-1}, t_m],
\end{equation}
where $\bm{g}_m$ (parameterized by $\bm{\theta}_m$) governs the continuous transformation of the feature state, capturing gradual feature refinement and integration. The solution at the end of each interval, is denoted $\bm{X}_m = \bm{h}(t_m)$. We approximate $\bm{g}_m$ by a truncated Volterra filter. 



\item \textbf{Classification Layer}: At each stage $m$, we add a classification layer to help with optimization of extracted features using cross entropy. This enables stable training of the model by guiding each stage to learn relevant features. The classification layer computes the decision based on $\bm{X}_m$ from the ODE Block, 
\begin{equation}
    \bm{\hat{y}}_m = \bm{c}_m(\bm{X}_m) =  \sigma \Bigg( \bm{W}_{m} \text{ flatten}(\bm{X}_m) \Bigg) + b_m,
\end{equation}
where, $\sigma$ is the softmax layer, $\bm{W}_m$ are the classifier weights for the $m^{th}$ stage, and $b_m$ is the bias. At the final layer, $\bm{\hat{y}}_M$ is treated as the final decision of the model. The block diagram summarizing our approach is depicted in Figure \ref{fig:bd}. 
\end{enumerate}


\subsection{Loss Function and Training}

Like in \cite{chen2018neural}, we treat the ODE solver as a black box and compute the gradients using adjoint sensitivity method \cite{pontryagin2018mathematical}.  This approach scales linearly with problem size, has low memory cost, and explicitly controls numerical error. Consider a loss function $\bm{\mathcal{L}}_m(\cdot)$ at stage $m$, whose input is the result of an ODE slover, 

    \begin{align}
    &\bm{\mathcal{L}}_m(\bm{\hat{y}}_m, \bm{y}_{gt}) = \bm{\mathcal{L}}_m (\bm{c}_m(X_m), \bm{y}_{gt}) \\ &= \bm{\mathcal{L}}_m \Bigg(\bm{c}_m\bigg( \bm{h}(t_{m-1}) + \int_{t_{m-1}}^{t_m} \bm{g}_m(\bm{h}(t), t; \bm{\theta}_m) dt \bigg), \bm{y}_{gt}\Bigg) \\
    &= \small\bm{\mathcal{L}}_m \Bigg( \bm{c}_m \bigg( \text{ODESOLVE} \big( \bm{h}(t_{m-1}), \bm{g}_m, t_{m-1}, t_m; \bm{\theta}_m \big) \bigg), \bm{y}_{gt} \Bigg).
    \end{align}

The overall cost function to be optimized becomes, 
\begin{equation}
    \min_{\bm{\Theta}, \bm{W}, \bm{V}}\sum_{m=1}^M \mathcal{L}_m (\hat{y}_m, y_{gt}; \bm{\theta}_m, \bm{W}_m, \bm{V}_m), 
\end{equation}
where, $\bm{\Theta} = \{ \bm{\theta}_m \}_{m=1}^M$ are the weights of the continuous Volterra filter $\{ \bm{g}_m \}_{m=1}^M$, $\bm{V} = \{ \bm{V}_m \}_{m=1}^M$ are the weights of the discrete Volterra filter, and $\bm{W} = \{ \bm{W}_m \}_{m=1}^M $ are the weights of the classifier layer. Each $\bm{\mathcal{L}}_m$ in the above equation is a cross-entropy loss applied to $\bm{\hat{y}}_m$,

\begin{equation}
    \bm{\mathcal{L}}(\bm{\hat{y}}, \bm{y}) = \sum_i -y_i \log(\hat{y}_i).
\end{equation}

\section{Experiments and Results}
 \begin{table*}[]
        \centering
        \begin{tabular}{c|c|c|c|c}
        \hline
            Type & Model & Params (M) & GFLOPs & ImageNet-1K\\
            \hline
             CNN & VGG-16 \cite{simonyan2014very} & 138 & 15.4 & 75.2 \\
             CNN & DenseNet-121 \cite{huang2017densely} & 8 & 5.5 & 74.98 \\
             CNN & ResNet-50 \cite{he2016deep} & 25.6 & 4.09 & 76.1 \\
             CNN & ResNet-101 \cite{he2016deep} & 44.5 & 7.8 & 77.37 \\
             CNN & ResNet-152 \cite{he2016deep} & 60.2 & 11.51 & 78.31 \\
             CNN & Inception-ResNet-v2 \cite{szegedy2017inception} & 54.8 & 15.1 & 80.1 \\
             CNN & MobileNet-v3 & 5.4 & 0.3 & 75.2\\
             CNN & ConvNeXt-T \cite{liu2022convnet} & 29 & 4.5 & 82.1 \\
             CNN & E-ConvNeXt-Mini \cite{wang2025convnext} & 7.6 & 0.93 & 78.3 \\
             CNN & E-ConvNeXt-Tiny \cite{wang2025convnext} & 13.2 & 2.04 & 80.6 \\
             CNN & E-ConvNext-Small \cite{wang2025convnext} & 19.4 & 3.12 & 81.9\\
             \hline
             Transformers & TinyViT  \cite{wu2022tinyvit} & 21 & 4.4 & 83.1 \\
             \hline
             Hybrid (CNN+Trans) & MobileVIT-XS \cite{mehta2021mobilevit} & 2.3 & 0.7 & 74.8 \\
             \hline
             VNN & Vanilla VNN  \cite{roheda2020conquering} & 12 & 3.6 & \underline{83.3} \\
             \hline
             ODE Based  & NODE \cite{chen2018neural} & 0.7 & 0.3 & 68.3 \\
             ODE Based & MALI \cite{zhuang2021mali}& 11.2 & - & 70.17 \\
             \hline
             Continuous VNN  & \textbf{VNODE} (M=1) & 1.5 & 0.4 & 78.6 \\
             \hline
             \makecell{Piecewise \\ Continuous VNN} & \textbf{VNODE} (M=3)  & 4.5 & 1.2 & 81.1\\
              & \textbf{VNODE} (M=6)  & 9.1 & 2.4 & \textbf{83.5}\\
             \hline
             
        \end{tabular}
        \caption{Quantitative results on and ImageNet-1k.}
        \label{tab:C10_IN1K}
    \end{table*}

\begin{table}[]
    \centering
    \begin{tabular}{c|c|c|c}
        \hline
        \small Model & \small Params (M) & \small GFLOPs & \small Acc (\%) \\
        \hline
        \small ResNet-110 \cite{he2016deep} & \small 1.7 & \small 0.5 & \small 93.57 \\
        \small DenseNet-BC \cite{huang2017densely} & \small 15.3 & \small - & \small 94.81 \\
        \small Vanilla VNN \cite{roheda2020conquering} & \small 1.5 & \small 0.65 & \small 94.2 \\
        \small NODE \cite{chen2018neural} & \small 0.7 & \small 0.3 & \small 84.62 \\
        \small MALI \cite{zhuang2021mali} & \small 11.2 & \small - & \small 93.7 \\
        \hline
        \small \textbf{VNODE} (M=1) & \small 0.5 & \small 0.11 & \small 89.6 \\
        \small \textbf{VNODE} (M=3) & \small 0.9 & \small 0.33 & \small \underline{94.90} \\
        \small \textbf{VNODE} (M=4) & \small 1.2 & \small 0.45 & \small \textbf{95.1} \\

    \end{tabular}
    \caption{Performance comparison on CIFAR-10 dataset}
    \label{tab:cifar}
\end{table}
    

\subsection{Image Classification}
\begin{figure}[b]
    \centering
    \includegraphics[width=\linewidth]{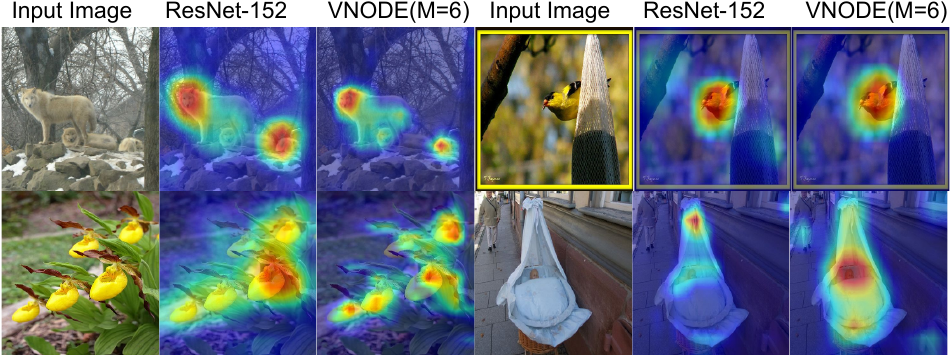}
    \caption{Comparison of Gradcam on ResNet152 and VNODE}
    \label{fig:gradcam}
\end{figure}
We evaluate the proposed method on standard benchmark datasets, including CIFAR-10 \cite{krizhevsky2009learning} and ImageNet-1K \cite{deng2009imagenet}. In our architecture, feature extraction at each stage is performed using a second-order Volterra filter. The continuous evolution of features is modeled via an ordinary differential equation (ODE), where the function $\bm{g}_m(\cdot)$ is also instantiated using a second-order Volterra filter. Our approach introduces a key modification relative to the NODE baseline \cite{chen2018neural}. Rather than concatenating the temporal variable $t$ with the hidden state $\bm{h}(t)$, we apply element-wise addition, resulting in the input to $\bm{g}_m(\cdot)$ being $\bm{h}(t)+t$. This design choice enables the use of groups within $\bm{g}_m(\cdot)$, thereby reducing the overall model complexity. The volterra filter is implemented using a lossy approximation as detailed in \cite{roheda2024mr}.

\textbf{CIFAR-10}: For CIFAR-10, we employ an architecture comprising of $3 \times 3$ Volterra convolutional kernels. The initial discrete filter expands the input channel dimension from $3$ to $64$, after which the channel count remains fixed throughout the network. When the model is configured with a single stage ($M = 1$), the resulting architecture corresponds to a fully continuous VNN. In Table \ref{tab:cifar}, we present a comparative analysis of our model against SOTA methods. The results demonstrate that VNODE achieves superior performance while significantly reducing model complexity. 

\textbf{ImageNet-1K}: To further validate the effectiveness of our proposed method, we conduct experiments on the ImageNet-1K dataset. The initial discrete Volterra filter employs a $7 \times 7$ Volterra convolutional kernel to extract $64$ feature channels. Subsequent stages adopt a multi-scale feature extraction strategy inspired by the Inception architecture \cite{szegedy2017inception}, utilizing parallel Volterra filters with kernel sizes of $1 \times 1$, $3 \times 3$, and $5 \times 5$.
The network is composed of six such stages, progressively increasing the feature dimensionality to $1024$ channels. This hierarchical design enables efficient representation learning while maintaining computational tractability.
Table \ref{tab:C10_IN1K} presents a quantitative comparison between the proposed VNODE model and leading SOTA methods. The results demonstrate that using a piecewise continuous formulation results in higher performance and lower model complexity. Additionally, Figure \ref{fig:gradcam} shows GradCam \cite{selvaraju2017grad} visualizations for ResNet-152 and VNODE. These visualizations indicate that our model achieves more accurate localization of the object of interest and reduces the influence of background regions in its predictions. Figure \ref{fig:gflopsvsacc} summarizes model accuracy versus computational complexity.


\subsection{Robustness analysis}
To demonstrate the robustness of the proposed model, we also show results on the CIFAR-10C dataset \cite{hendrycks2018benchmarking} which includes 15 types of image corruptions (eg. noise, blur, etc.) at 5 severity levels. Table \ref{tab:cifar-10c} showcases the impressive robustness of our model, without requiring any additional augmentations or fine-tuning. The accuracy is reported as an average over all the 75 corruptions in the CIFAR-10c dataset.

\begin{table}[]
    \centering
    \begin{tabular}{c|c|c}
    \hline
         \small Model & \small CIFAR-10 (\%) & \small CIFAR-10C (\%) \\
         \hline
         \small ResNet-110 & \small 93.57 & \small \underline{74.2}\\
         \small DenseNet-BC & \small \underline{94.81} & \small 67.2\\
         \small \textbf{VNODE} & \small \textbf{95.1} & \small \textbf{78.9}\\
    \end{tabular}
    \caption{Robustness evaluation on CIFAR-10C dataset}
    \label{tab:cifar-10c}
\end{table}


\section{Conclusion}
The proposed VNODE unites nonlinear Volterra filtering with continuous-time dynamics, capturing complex visual patterns through alternating discrete and continuous processing, mirroring principles from the visual cortex. Experiments on major benchmarks show that VNODE surpasses SOTA methods while requiring fewer parameters and lower computational cost than standard CNNs or transformers. 

\small\bibliographystyle{IEEEbib}
\bibliography{strings,refs}

\end{document}